\documentclass[runningheads]{llncs}
\usepackage{graphicx}
\usepackage{comment}
\usepackage{amsmath,amssymb} 
\usepackage{color}
\usepackage[font=small]{caption}

\usepackage{multirow}
\usepackage{graphicx}
\usepackage[table,xcdraw,dvipsnames]{xcolor}
\usepackage{alex}
\usepackage{epsfig}
\usepackage{booktabs}
\usepackage{xr}
\usepackage[normalem]{ulem}
\usepackage{multirow}
\usepackage{enumitem}
\usepackage{subfigure}
\usepackage[symbol]{footmisc}
\usepackage{caption}
\usepackage{url}
\usepackage[bookmarks=false]{hyperref}

\newenvironment{packed_enum}{
\begin{enumerate}
  \setlength{\itemsep}{2.5pt}
  \setlength{\parskip}{0pt}
  \setlength{\parsep}{0pt}
}{\end{enumerate}}
\newcommand{\mypara}[1]{\vspace{2pt}\noindent\textbf{#1}}

\newcommand{\KGcr}[1]{{\color{black}{}#1}}  

\newcommand{\KG}[1]{{\color{black}{}#1}} 
\newcommand{\K}[1]{{\color{black}{}#1}}
\newcommand{\cc}[1]{{\color{black}{}#1}} 
\newcommand{\unnat}[1]{{\color{black}{}#1}} 

\newcommand{\C}[1]{{\color{black}{}#1}}

\newcommand{\cca}[1]{{\color{black}{}#1}} 
\newcommand{\ccan}[1]{{\color{black}{}#1}} 
\newcommand{\cn}[1]{{\color{black}{}#1}} 

\newcommand{\ZA}[1]{{\color{black}{}#1}} 
\newcommand{\Z}[1]{{\color{black}{}#1}} 
\newcommand{\Zcr}[1]{{\color{black}{}#1}} 

\begin{document}

\pagestyle{headings}
\mainmatter
\def\ECCVSubNumber{4577}  


\title{SoundSpaces: Audio-Visual Navigation \\ in 3D Environments}
\author{%
Changan Chen\thanks{CC and UJ contributed equally;  $^{\dagger}$work done as an intern at Facebook AI Research}$^{1,4}$, Unnat Jain$^{*\dagger2,4}$,  Carl Schissler$^{3}$, Sebastia Vicenc \\Amengual Gari$^{3}$, Ziad Al-Halah$^{1}$,
Vamsi Krishna Ithapu$^{3}$, \\
Philip Robinson$^{3}$, and Kristen Grauman$^{1,4}$\\
}
\institute{
$^1$UT Austin, $^2$UIUC, $^3$Facebook Reality Labs, $^4$Facebook AI Research
}

\authorrunning{C. Chen \& U. Jain et al.}
\titlerunning{SoundSpaces: Audio-Visual Navigation  in 3D Environments}
\maketitle

\begin{abstract}

Moving around in the world is naturally a multisensory experience, but today's embodied agents are deaf---restricted to solely their visual perception of the environment.  We introduce audio-visual navigation for complex, acoustically and visually realistic 3D environments.  
By both seeing and hearing, the agent must learn to navigate to a sounding object.  
We \Z{propose} a multi-modal deep reinforcement learning \Z{approach} to train navigation policies end-to-end from a stream of egocentric audio-visual observations, allowing the agent to (1) discover elements of the geometry of the physical space indicated by the reverberating audio and (2) detect and follow sound-emitting targets.  
We further introduce \cn{SoundSpaces:} \KGcr{a first-of-its-kind dataset of}
audio renderings based on geometrical acoustic simulations for \KGcr{two sets} of publicly available 3D environments \KGcr{(Matterport3D and Replica)}, and we instrument Habitat to support the new sensor, making it possible to insert arbitrary sound sources in an array of \KGcr{real-world scanned} environments.   
Our results show that audio greatly benefits embodied visual navigation in 3D spaces, \KG{and our work lays groundwork for new research in embodied AI with audio-visual perception.}
\cca{\Zcr{Project}: \url{http://vision.cs.utexas.edu/projects/audio_visual_navigation}.}

\end{abstract}
\section{Introduction}

\K{Embodied agents perceive and act in the world around them, with a constant loop between their sensed surroundings and their selected movements.
Both sights and sounds constantly drive our activity: the laundry machine buzzes to indicate it is done, a crying child draws our attention, the sound of breaking glass may require urgent help.}

\begin{figure}[t]
    \centering
    \includegraphics[trim=0 0 0 0 ,clip,width=0.7\linewidth]{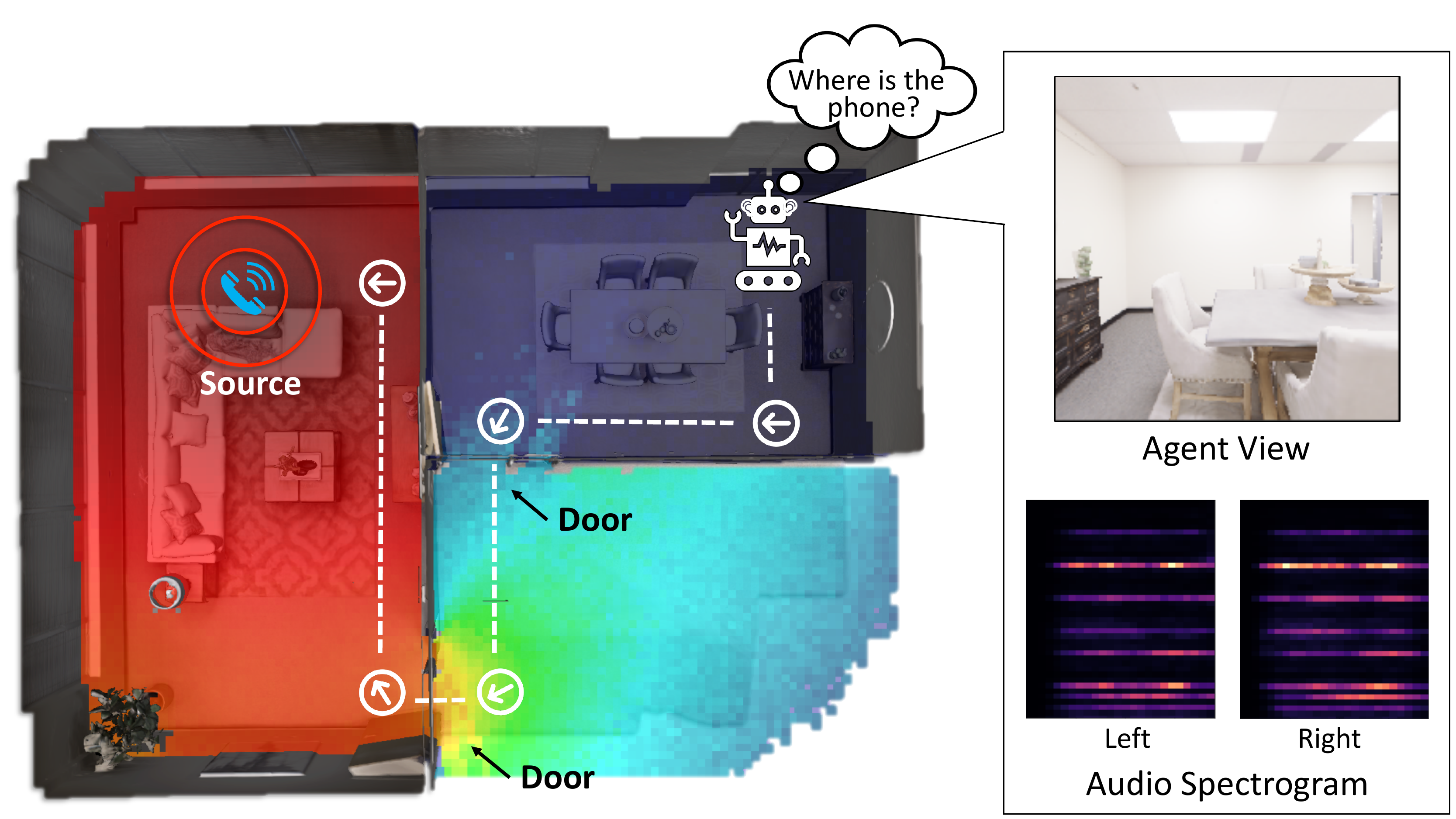}
    \caption{\small{\Z{\textbf{Audio source in an \KG{unmapped} 3D environment}, where an autonomous agent must navigate to the goal. The top-down map is overlaid with the acoustic pressure field heatmap.}
    Our audio-enabled agent gets rich directional information about the goal, since the \Z{audio intensity} variation is correlated with the shortest path distance.  The acoustics also reveal the room's geometry, major structures, and materials.
    \cc{Notice the gradient of the field along the \emph{geodesic} path an agent must use to reach the goal (different from the shortest Euclidean path, which would cut through the inner wall). As a result,} \KG{the proposed agent enjoys the synergy of both modalities:}
    audio reveals the door as a good intermediate goal, \KG{while vision reveals the physical obstacles along the path, such as the furniture \cc{in the lefthand room.}}}
    }
    \label{fig:concept}
\end{figure}

\K{In embodied AI, the \emph{navigation} task is of particular importance, with applications in search and rescue or service robotics, among many others.
Navigation has a long history in robotics, where a premium is placed on rigorous geometric maps~\cite{thrun,hartley}.  More recently, researchers in computer vision are exploring models that loosen the metricity of maps in favor of end-to-end policy learning and learned spatial memories that can generalize to visual cues in novel environments~\cite{ZhuARXIV2017,gupta2017unifying,gupta2017cognitive,savinov2018semiparametric,anderson-eval,mishkin2019benchmarking,habitat19iccv}.}

\K{However, while current navigation models tightly integrate seeing and moving, they are deaf to the world around them.  This poses a significant sensory hardship: sound is key to (1) understanding a physical space and (2) localizing sound-emitting targets.  As leveraged by blind people and animals who perform sonic navigation, acoustic feedback partially reveals the geometry of a space, the presence of occluding objects, and the materials of major surfaces~\cite{picinali,evers}---all of which can complement the visual stream.  Meanwhile, targets currently outside the visual range may be detectable \emph{only} by their sound (e.g., a person calling from upstairs, the ringing phone occluded by the sofa, \cc{footsteps approaching from behind}).  Finally, aural cues become critical when visual cues are  unreliable (e.g., the lights flicker off) or orthogonal to the agent's task (e.g., a rescue site with rubble that breaks prior visual context).} 

\K{Motivated by these factors, we introduce \emph{audio-visual navigation} for complex, visually realistic 3D environments.  The \KG{autonomous} agent can both see and hear while attempting to reach its target.  We consider two variants of the navigation task: (1) \emph{AudioGoal}, where the target is indicated by the sound it emits, and (2) \emph{\ZA{AudioPointGoal}}, where the agent is additionally directed towards the goal location at the onset.
The former captures scenarios where a target initially out of view makes itself known aurally (e.g., \KG{phone ringing}). 
The latter augments the \KG{popular}
PointGoal navigation task~\cite{anderson-eval} and captures scenarios where the agent has a GPS pointer towards the target, but should leverage audio-visual cues to navigate the unfamiliar environment and reach it faster.}

\K{We \Z{propose} a multi-modal deep reinforcement learning (RL) \Z{approach} to train navigation policies end-to-end from a stream of audio-visual observations.  Importantly, audio observations must be
generated with respect to both the agent's current position and orientation as well as the physical properties of the 3D environment.  To do so, we introduce pre-computed audio renderings \cn{SoundSpaces} for \KGcr{Matterport3D~\cite{chang2017matterport} and} Replica~\cite{straub2019replica}, \KGcr{two} public datasets of \KG{scanned real-world} 3D environments, and we integrate them with the open source Habitat platform~\cite{habitat19iccv} for fast 3D simulation (essential for scalable RL).  The proposed embodied AI agent learns a policy to choose motions in a novel, unmapped environment that will bring it efficiently to the target while discovering relevant aspects of the latent environment map.} See Figure~\ref{fig:concept}.

\K{Our results show 
\KG{the powerful synergy between} audio and vision for navigation.
The agent learns to blend both modalities to map \KGcr{novel} environments,
and doing so yields faster learning \cc{at training time} and faster, more accurate navigation \cc{at inference time}.  Furthermore\KG{---in one of our most exciting results---}we demonstrate that for an audio goal, the audio stream competes well with the goal displacement vectors \KG{upon which current navigation methods often depend}~\cite{anderson-eval,habitat19iccv,gordon2019splitnet,kojima2019learn,chaplot2020Explore}, while having the advantage of not assuming perfect GPS odometry.  Finally, we explore the agent's ability to generalize to not only unseen environments, but also unheard sounds.
}\K{Our main contributions are:
\begin{packed_enum}
    \item We introduce the task of audio-visual navigation \Z{by autonomous agents} in complex, visually \KGcr{and acoustically} realistic 3D environments.
    \item We generalize a state-of-the-art deep RL visual navigation \Z{framework} to accommodate audio observations and demonstrate its impact on navigation.
    \item \KGcr{We introduce \cn{SoundSpaces}, a first-of-its-kind audio-visual platform  for embodied AI.}  We instrument the \KGcr{103 environments from Matterport3D}~\cite{chang2017matterport} and Replica~\cite{straub2019replica} on the Habitat platform~\cite{habitat19iccv} with acoustically \Z{realistic} sound renderings.  This allows insertion of an arbitrary sound source and proper sensing of it from arbitrary agent receiver positions.  \KG{By sharing this new resource publicly, our work can enable other new ideas in this area.}
    \item We create a benchmark suite of tasks for audio-visual navigation to facilitate future work in this direction.
\end{packed_enum}
}

\section{Related Work}

\noindent{\textbf{Audio-visual learning.}}
\K{The recent surge of research in audio-visual \KGcr{(AV)} learning focuses on video rather than embodied perception.}  This includes interesting directions for synthesizing sounds for video~\cite{owens2016visually,chen2017deep,zhou2018visual}, \K{spatializing sound}~\cite{morgado2018self,gao201925Sound}, 
sound source separation~\cite{zhao2018sound,gao2018learning,owens2018audio,ephrat2018looking,gao2019co}, cross-modal feature learning~\cite{Aytar16,Aytar17,owens2016ambient,ruohan-eccv2020}, \KGcr{AV} tracking~\cite{gebru2015iccvw,ban2017iccvw,ban2018icassp,alameda2015salsa}, \K{and learning material properties~\cite{owens2016visually}.}
\K{Unlike prior work that localizes pixels in video frames associated with sounds~\cite{tian2018audio,senocak2018learning,arandjelovic2018objects,hershey2000audio}, our goal is to learn navigation policies \KG{for agents to} actively locate an audio target in a 3D environment.}
 \K{Unlike any of the above, our work addresses embodied navigation, not learning from human-captured video.}

\vspace*{0.1in}
\noindent{\textbf{Vision-based navigation.}}
The role of vision for cognitive mapping in \emph{human} navigation is well studied in neuroscience~\cite{ekstrom2015vision,tolman1948cognitive}.
\KGcr{Recent} AI agents also
\K{aggregate} egocentric visual inputs
\cite{ZhuARXIV2016,ZhuARXIV2017,mirowski2017learning,thomason2019shifting,JainWeihs2020CordialSync}, \K{often with a spatio-temporal memory}~\cite{gupta2017cognitive,savinov2018semiparametric,henriques2018mapnet,wu2019bayesian}. 
\cc{Visual navigation can be tied to other tasks to attain intelligent behavior,
\K{such as} question answering~\cite{GordonCVPR2018,DasECCV2018,das2020probing}, 
active visual recognition~\cite{jayaraman2018end},
and instruction following~\cite{anderson2018vision,chen2019touchdown}.} 
Our work goes beyond visual perception to incorporate hearing, offering a novel  perspective on  navigation.

\vspace*{0.1in}
\noindent{\textbf{Audio-based navigation.}}
Cognitive science also confirms that audio is a strong navigational signal~\cite{thinus1997representation,merabet2010neural}.
\cc{Blind and sighted people
show comparable skill on 
spatial navigation~\cite{fortin2008wayfinding} and sound localization~\cite{gougoux2005functional,lessard1998early,roeder1999improved,voss2004early} tasks.
Consequently, audio-based AR/VR equipment has been devised for auditory sensory substitution for human users for obstacle avoidance and navigation~\cite{massiceti2018stereosonic,gunther_using_2004}.  
Additionally,} \KG{cartoon}-like virtual 2D and 3D AV environments  \K{can help evaluate human} learning of audio cues~\cite{connors2013development,wood2003design,merabet2009audio}.
\ZA{Unlike our proposed platform,} these environments are non-photorealistic and \KG{they are for \emph{human} navigators}; they do not support AI agents or training.
 \K{Prior studies with autonomous agents in simulated environments are restricted to human-constructed game boards, do not use acoustically correct sound models, 
 and train and test on the same environment~\cite{wang2014,woubie2019autonomous}.}
 
\vspace*{0.1in}
\noindent{\textbf{Sound localization in robotics.}}
In robotics, microphone arrays are often used for sound source localization~\cite{nakadai1999sound,rascon2017localization,nakadai2000active,nakadai2001epipolar}. 
Past studies fuse \KGcr{AV} cues for surveillance~\cite{wu2009surveillance,qin2006learning}, speech recognition~\cite{yoshida2009automatic}, human robot interaction~\cite{alameda2015vision,viciana2014audio}, and robotic manipulation tasks~\cite{romano2013ros}.  
\KG{None attempt audio-visual navigation in unmapped environments.}
\KGcr{Concurrent work explores AV-navigation in computer graphics environments~\cite{gan-icra2020}.  In contrast to our end-to-end RL agent, their model decouples the task into predicting the goal location from audio and then planning a path to it.  Our simulation platform is more realistic for both visuals (real world images in ours vs.~computer graphics in~\cite{gan-icra2020}) and acoustics (ray tracing/sound penetration/full occlusion model in ours vs.~low-cost game audio in~\cite{gan-icra2020}), and it offers 
5,000$\times$ more audio data and 15$\times$ more environments.}   
\K{To our knowledge, ours is the first work to demonstrate improved navigation by an AV agent in a visually and acoustically realistic 3D environment, and the first to introduce an end-to-end approach for the problem.}

\vspace*{0.1in}
\noindent{\textbf{3D environments.}}
\K{Recent research in embodied perception is greatly facilitated by new 3D environments and simulation platforms. 
Compared to artificial environments like video games~\cite{kempka2016vizdoom,lerer2016learning,johnson2016the,wymann2013torcs,SukhbaatarARXIV2015}, 
photorealistic environments portray 3D scenes in which real people  \KG{and mobile robots} would interact.   
Their realistic meshes can be rendered from agent-selected viewpoints to train and test RL policies for navigation in a reproducible manner~\cite{ammirato2016avd,chang2017matterport,xia2018gibson,ai2thor,stanford2d3d,straub2019replica,BrodeurARXIV2017,xia2019interactive,habitat19iccv}.}  \KG{Many are captured with 3D scanners and real 360 photos, meaning that the views are indeed the perceptual inputs a robot would receive in the real world~\cite{chang2017matterport,straub2019replica,ammirato2016avd}.}
None of the commonly used environments and simulators provide audio \KGcr{rendering}. 
We present the first audio-visual simulator for AI agent training and \K{the first study of audio-visual embodied agents in realistic 3D environments.}

\label{sec:platform}
\begin{figure}[t]
    \centering
    \begin{tabular}{c}
    \includegraphics[trim=0 0 0 0,clip,width=0.45\linewidth]{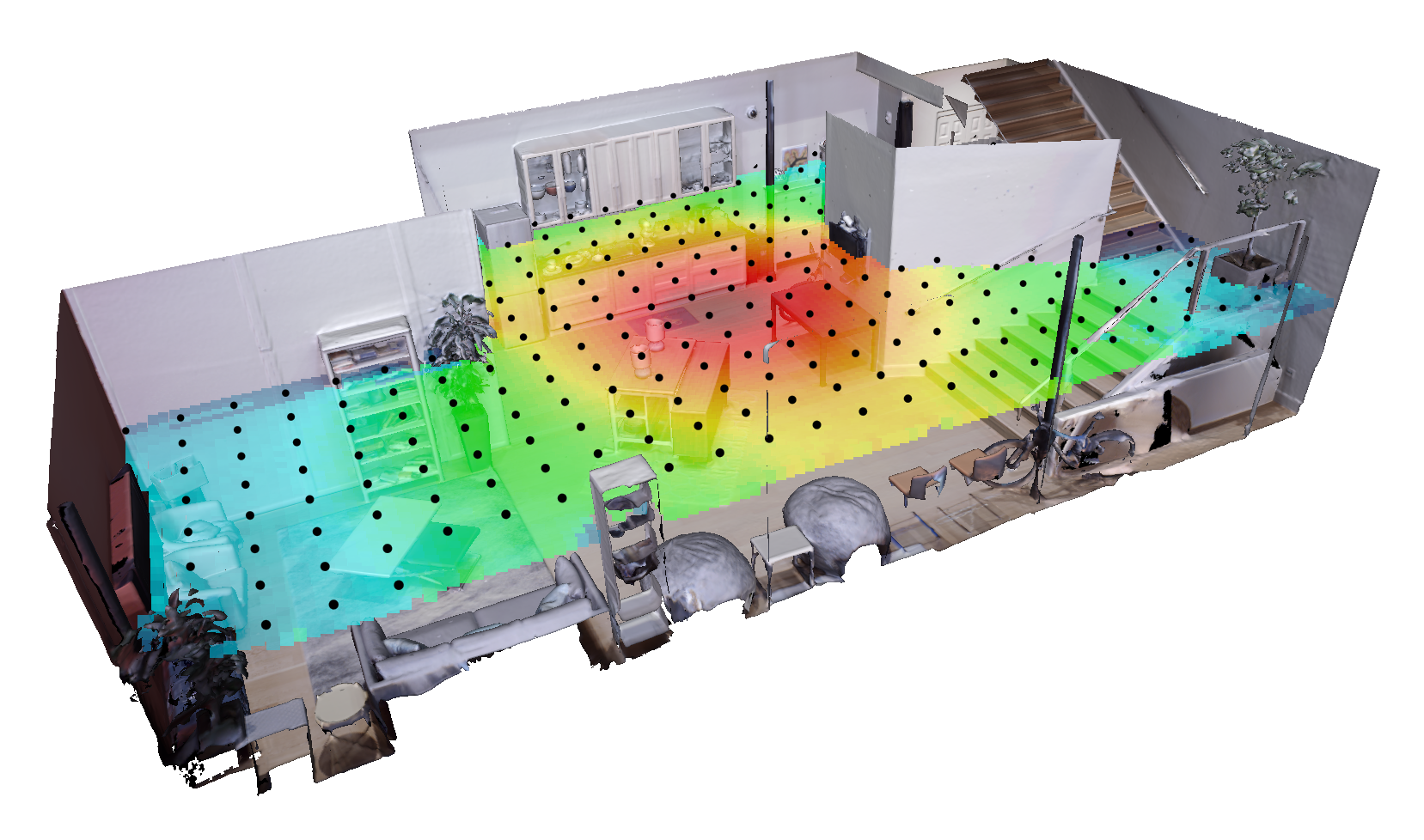}
    \includegraphics[width=0.45\linewidth]{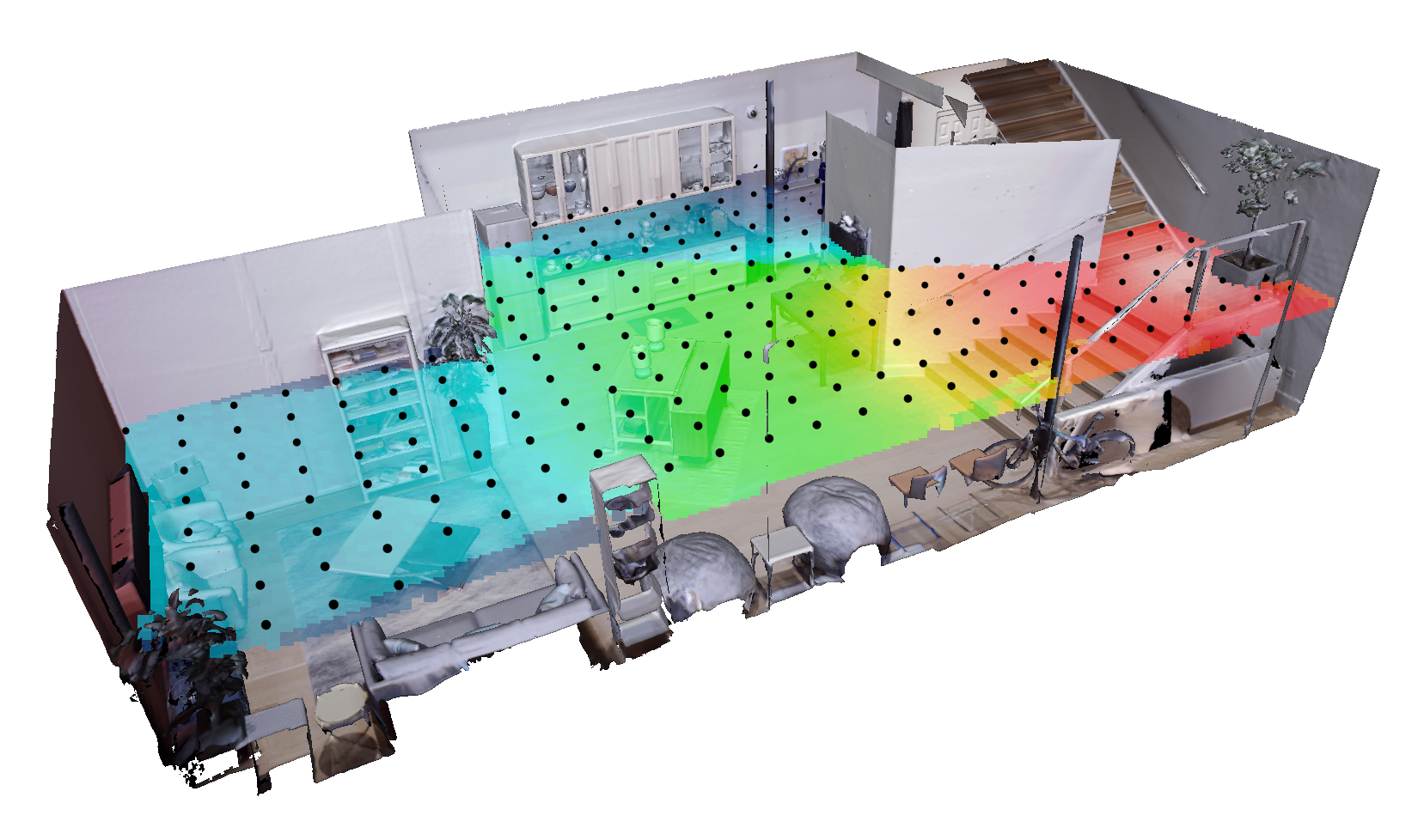}
    \end{tabular}
    \vspace*{-0.1in}
    \caption{\small{\textbf{Acoustic simulation.} We capture room impulse responses between each location pair within the illustrated grid (here for the `frl\_apartment\_0' scene in Replica).
    In our platform, agents can experience binaural audio at densely sampled locations $\mathcal{L}$ marked with black dots---hearing the sound's intensity, direction, and frequency texture.  Heatmaps display audio pressure fields, decreasing from red to blue. \textbf{Left}: When a sound source in $\mathcal{S}$ is placed in the center. \textbf{Right}: When a source is placed on the stairs. Notice how the sound received by the agent at different positions changes when the sound source moves,
    and how 3D structures influence the sound propagation.}}\label{fig:example-replica}
    \vspace*{-0.18in}
\end{figure}

\section{\cn{SoundSpaces: Enabling Audio in Habitat}}
\vspace*{-0.1in}

Our audio platform augments the Habitat simulator~\cite{habitat19iccv}, particularly the \KGcr{Matterport3D~\cite{chang2017matterport} and} Replica~\cite{straub2019replica} datasets hosted within it.
Habitat is an open-source 3D simulator with a user-friendly API \ZA{that}
supports RGB, depth, and semantic rendering.
The API \K{offers} fast (over 10K fps) rendering and support for multiple datasets~\cite{straub2019replica,xia2018gibson,Chang3DV2017Matterport,pyrobot2019,DasCVPR2018}.
This has incentivized many embodied AI works to embrace it as the 3D simulator for training navigation and question answering agents~\cite{habitat19iccv,chaplot2020Explore,kojima2019learn,gordon2019splitnet,Wijmans2019EQAPhoto}.

We use 85 \KGcr{Matterport3D~\cite{chang2017matterport} environments, which are real-world homes and other indoor environments with 3D meshes and image scans.
The environments are large, with on average 517 m$^2$ of floor space.}
Replica~\cite{straub2019replica} is a dataset of 18 apartment, hotel, office, and room scenes with 3D meshes.
By extending \KGcr{these Habitat-compatible 3D assets} with our audio simulator, we enable users to take advantage of the efficient Habitat API and easily adopt the audio modality for AI agent training. 
Our audio platform and data \KGcr{is} shared publicly.

Our \KG{high-fidelity} audio simulator \cn{SoundSpaces} takes into account important factors for a realistic sound rendering in a 3D environment.
We use a state-of-the-art algorithm for room acoustics modeling~\cite{cao2016} and a bidirectional path tracing algorithm to model sound reflections \KGcr{in the room geometry}~\cite{veach1995}.
\KGcr{Since} materials also influence the sounds received in an environment (e.g., walking across marble floors versus a shaggy carpet), 
we set the acoustic material properties of \KG{major surfaces}
\KGcr{by mapping the meshes' semantic labels to materials in an existing database~\cite{egan1989}.}
Each material has different absorption, scattering, and transmission coefficients \KG{that affect our sound propagation} \KG{(see Supp)}.
This enables our simulator to model fine-grained acoustic properties like sound propagation through walls.

For each scene, we simulate the acoustics of the \K{environment by pre-computing room impulse responses (RIR).
The RIR is the transfer function between a sound source and microphone, which varies as a function of the room geometry, materials, and the sound source location~\cite{kuttruff2016}.}

Let $\mathcal{S}=\{(x_i^s,y_i^s,z_i^s)\}_{i=1}^N$ denote the set of $N$ possible sound source positions, and let $\mathcal{L} = \{(x_i^r,y_i^r,z_i^r)\}_{i=1}^N$ denote the set of possible listener positions (i.e., agent microphones).
We densely sample a grid of $N$ locations with spatial resolution of $0.5$m \KGcr{(Replica) or 1m (Matterport)}.  
\KG{The Replica scenes range in area from 9.5 to 141.5 m$^2$ and thus yield $N \in [38,566]$;  \KGcr{for Matterport the range is 53.1 to 2921.3 m$^2$, with $N \in [20,2103]$.}}
\KG{Points are placed at a vertical height of $1.5$m, reflecting the fixed height of a robotic agent.} 
Then we simulate the RIR \K{for each possible} source and \K{listener placement} at these locations, \K{$\mathcal{S} \times \mathcal{L}$.
Having done so, we can look up any source-listener pair on-the-fly and render the sound, by convolving the desired waveform with the selected RIR.}  See Figure~\ref{fig:example-replica}. 

\KGcr{Given our simulations}, for any audio source placed in a location $\mathcal{S}_i$ we can generate the ambisonic audio (roughly speaking, the audio equivalent of a 360$^\circ$ image) heard at a particular listener location $\mathcal{L}_j$.
We convert the ambisonics to binaural audio~\cite{zaunschirm2018binaural} in order to represent an agent with two human-like ears, for whom perceived sound depends on the body's relative orientation in the scene.\footnote{\KG{While algorithms could also run with ambisonic inputs, using binaural sound has the advantage of allowing human listeners to interpret our video results (see Supp video).}}  \KGcr{Our platform also permits rendering multiple simultaneous sounds.}

\KG{Since} an agent might not be able to stand at each location in $\mathcal{L}$ due to embodiment constraints \KG{(e.g., no climbing on the sofa)}, we create a graph capturing the reachability and connectivity of these locations.
\ZA{First we remove nodes that are} non-navigable, \ZA{then} for each node pair $(i, j)$, we consider \ZA{the edge} $e(i, j)$ as valid if and only if the Euclidean distance between $i$ and $j$ is $0.5$m for Replica or 1m for Matterport (\ie, \ZA{nodes $i$ and $j$ are} immediate neighbors) and the geodesic and Euclidean distances between them are
\Zcr{equal}
(\ie, no obstacle in between).

All details of our audio simulation are in the Supp.  The fidelity of the sound renderings can be experienced  \KGcr{in our project page videos}.
\section{\K{Task Definitions: Audio-Visual Navigation}}
\label{sec:task}

We propose two novel navigation tasks: AudioGoal Navigation and AudioPointGoal Navigation.
In AudioGoal, the agent hears an audio  source located at the goal---such as a phone ringing---but receives no direct position information about the goal.  AudioPointGoal is an  audio extension of the PointGoal task studied often in the literature
\cite{anderson-eval,habitat19iccv,gordon2019splitnet,xia2019interactive,kojima2019learn,chaplot2020Explore} where the agent hears the source and is told its displacement from the starting position.
\KG{In all three tasks, to navigate and avoid obstacles, the agent needs to reach the target using sensory inputs alone.  That is, no map of the scene is provided to the agent.}

 \vspace*{-0.05in}
\paragraph{Task definitions.} For \ZA{PointGoal}~\cite{anderson-eval,habitat19iccv,wijmans2020ddppo}, a randomly initialized agent is tasked with navigating to a point goal defined by a 
\KGcr{displacement} vector $(\Delta_{x}^{0}, \Delta_{y}^{0})$ relative to the starting position of the agent.
For \K{AudioGoal}, the \KGcr{agent instead receives audio from the sounding target;}
\K{the AudioGoal agent does not receive} a displacement vector \K{pointing to the target}.
The observed audio is updated as a function of the location of the agent, the location of the goal, and the structure \K{and materials} of the room.
In {AudioPointGoal}, the agent receives the union of information received in the {PointGoal} and {AudioGoal} tasks, \cc{\ie, audio as well as a point vector}.
Note that physical obstacles (walls, furniture) typically exist along the displacement vector, which the agent must sense while navigating.

 \vspace*{-0.05in}
\paragraph{Agent and goal embodiment.} We adopt the standard cylinder embodiment used in Habitat.
A target has diameter $0.2$m and height $1.5$m, \K{and, consistent with prior \ZA{PointGoal} work, has no visual presence.}  While the goal itself does not have a visible embodiment \KG{(currently unsupported in Habitat)}, vision---particularly in the abstraction of depth---is \K{essential} to detect and avoid obstacles \K{to move towards the target}. Hence, 
\K{all} the tasks have a crucial vision component.

 \vspace*{-0.05in}
\paragraph{Action space.} The action space is: \emph{MoveForward}, \emph{TurnLeft}, \emph{TurnRight}, and \emph{Stop}. The last three actions are always valid. The \emph{MoveForward} action is invalid when the agent attempts to traverse from one node to another without an edge connecting them (as per the graph defined in~\secref{sec:platform}). If valid, \emph{MoveForward} takes the agent forward by $0.5$m \KGcr{(Replica) or 1m (Matterport).}
For all models, there is no actuation noise, \ie, a step executes perfectly or does not execute at all.

 \vspace*{-0.05in}
\paragraph{Sensors.} \K{The sensory inputs are binaural sound (absent in PointGoal), GPS \Zcr{(absent in AudioGoal)}, RGB, and depth. To capture binaural spatial sound, the agent emulates two microphones placed at human height.}  \K{We assume an idealized GPS sensor, following prior work
\cite{habitat19iccv,chaplot2020Explore,gordon2019splitnet,kojima2019learn}}.
\K{However, as we will demonstrate in results, our audio-based learning provides a steady navigation signal that makes it feasible to disable the GPS sensor for the proposed AudioGoal task.}

 \vspace*{-0.05in}
\paragraph{Episode specification.} An episode of \ZA{PointGoal} is defined by an arbitrary 1) scene,
2) agent start location, 3) agent start rotation, and 4) goal location.  In each episode the agent can reach the target \K{if it navigates successfully}.
An episode for AudioGoal and AudioPointGoal additionally includes a
source audio \K{waveform.} The \K{waveform} is convolved with the RIR corresponding to the specific scene, goal, agent location and orientation to generate dynamic audio for the agent. We consider a variety of audio sources, both familiar and unfamiliar to the agent \KGcr{(detailed below).}
An episode is successful if the agent executes the \emph{Stop} action while being exactly at the location of the goal. Agents are allowed a time horizon of $500$ actions for all tasks, similar to~\cite{habitat19iccv,jain2019two,chaplot2020Explore,gordon2019splitnet,kojima2019learn}.

\section{\K{Navigation Network and Training}}
\label{sec:approach}

\K{To navigate autonomously, the agent must be able to enter a new yet-unmapped space, accumulate partial observations of the environment over time, and efficiently transport itself to a goal location. \cc{Building on recent embodied visual navigation work~\cite{ZhuARXIV2017,gupta2017unifying,gupta2017cognitive,anderson-eval,mishkin2019benchmarking,habitat19iccv}}, we take a deep reinforcement learning approach, and we introduce audio to the observation.  During training, the agent is rewarded for correctly and efficiently navigating to the target.  This yields a policy  that maps new multisensory \KG{egocentric} observations to agent actions.}

\vspace*{-0.05in}
\paragraph{\K{Sensory inputs.}}
The audio inputs are \K{spectrograms}, following literature in audio \K{learning}~\cite{owens2016ambient,zhao2018sound,gao201925Sound}.
Specifically, to represent the agent's binaural audio input (corresponding to the left and right ear), \C{we first compute the Short-Time Fourier Transform (STFT) with a hop length of 160 samples and a windowed signal length of 512 samples, which corresponds to a physical duration of 12 and 32 milliseconds at a sample rate of 44100Hz (Replica) and 16000Hz (Matterport). By using the first 1000 milliseconds of audio as input,
STFT gives a $257\times257$ and a $257\times101$ complex-valued matrix, respectively; we take its magnitude and downsample both axes by a factor of $4$.} For better contrast we take its logarithm. Finally, we stack the left and right audio channel matrices to obtain a $65\times65\times2$ and a $65\times26\times2$ tensor, \K{denoted} $A$.
The visual input $V$ is the RGB and/or depth image,  $128\times128\times3$ and $128\times128\times1$  tensors, respectively, \K{where 128 is the image resolution for the agent's $90^\circ$ field of view.}
The relative displacement vector \K{$\Delta = (\Delta_{x}, \Delta_{y})$}
\K{points} from the agent to the goal in the 2D \K{ground} plane of the scene.

\begin{figure}[t]
    \centering
    \includegraphics[trim=0 0 0 0,clip,width=0.8\linewidth]{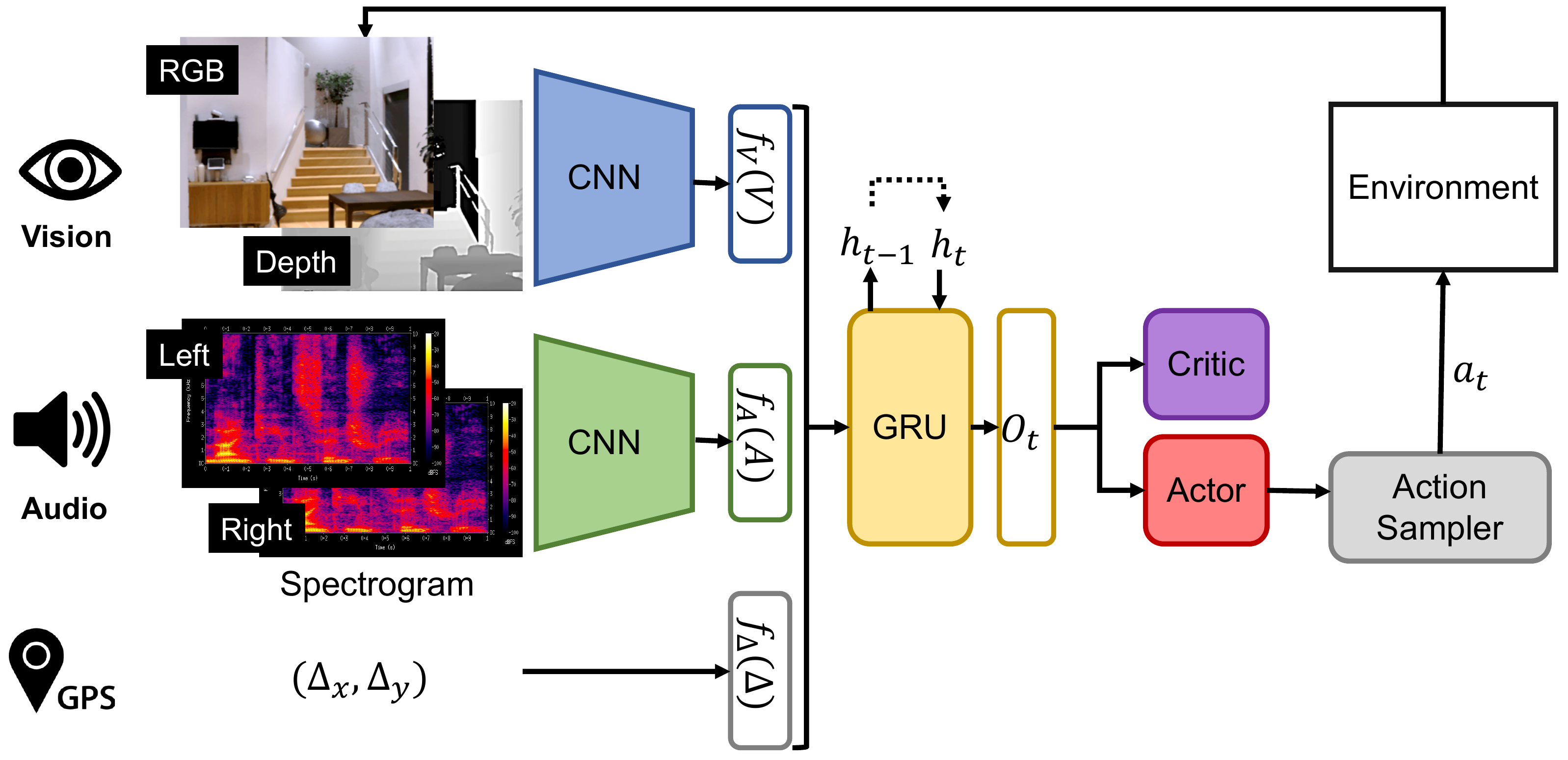}
    \vspace*{-0.1in}
    \caption{\small{\textbf{\Z{Audio-visual} navigation network.} \Z{Our model uses both acoustic and visual cues from the 3D environment for effective navigation of complex scenes.}}
    }
    \label{fig:model}
\end{figure}

Which specific subset of these three inputs \K{(audio, visual, vector)} the agent receives depends on the the agent's sensors and the goal's characterization \K{(cf.~Sec.~\ref{sec:task})}. 
The sensory inputs are transformed to a probability distribution over the action space by the policy network, as we describe next.

 \vspace*{-0.05in}
\paragraph{Network architecture.}
\unnat{Next we define the parameterization of the agent's policy \linebreak $\pi_\theta(a_t|o_t, h_{t-1})$, \K{which selects action $a_t$ given the current observation $o_t$ and aggregated past states $h_{t-1}$, and the} value \K{function} $V_\theta(\KGcr{o_t},h_{t-1})$, \K{which scores how good the current state is.}  Here $\theta$ \K{refers to} all trainable weights of the network.
}

\KG{Our network architecture is inspired by current RL models in the visual navigation literature~\cite{habitat19iccv,wortsman2019learning,DasCVPR2018,jain2019two}. We expand the traditional} vision-only navigation model to enable acoustic perception for audio-visual navigation.
As highlighted in~\figref{fig:model}, we transform $A$ and $V$ by corresponding CNNs $f_{A}(\cdot)$ and $f_{V}(\cdot)$. The CNNs have separate weights but the same architecture of conv $8\times8$, conv $4\times4$, conv $3\times3$ and a linear layer, with ReLU activations between each layer. The outputs of the CNNs are vectors $f_{A}(A)$ and $f_{V}(V)$ of length $L_A$ and $L_V$, respectively. These are concatenated to the relative displacement vector $\Delta$ and transformed by a \K{gated recurrent unit} (GRU)~\cite{ChungNIPS2015}. The GRU operates on the current step's input as well as the accumulated history of states $h_{t-1}$. The GRU updates the history to $h_t$ and outputs the representation of the agent's state $o_t$. 
Finally, the value of the state $V_\theta(o_t,h_{t-1})$
and the policy distribution $\pi_\theta(a_t|o_t, h_{t-1})$ are estimated using the critic and actor heads of the model. Both are linear layers.

\vspace*{-0.05in}
\paragraph{\unnat{Training.}} We train the network 
with Proximal Policy Optimization (PPO)~\cite{ppo}. 
\K{The agent is rewarded for reaching the goal quickly.  Specifically,}
it receives a reward of $+10$
for executing \textit{Stop} at the goal location, 
\K{a negative reward of $-0.01$ per time step}, $+1$ for reducing the geodesic distance to the goal, and the equivalent penalty for increasing it. 
\K{We add an entropy maximization term to the cumulative reward optimization, for better action space exploration~\cite{haarnoja2018soft,ppo}.}

 \vspace*{-0.05in}
\paragraph{\K{\KG{Synergy} of audio for navigation.}}
\K{Because our agent can both hear and see, it has the potential to not only better localize the target (which emits sound), but also better plan its movements in the environment (whose major structures, walls, furniture, etc.~all affect how the sound is perceived).  See Figure~\ref{fig:concept}.}
The optimal policy \K{would} trace a path $\mathcal{P}^{*}$ corresponding to monotonically decreasing geodesic distance \K{to the goal}. Notably, the displacement $\Delta$ does not
\K{specify} the optimal policy: moving along $\mathcal{P}^{*}$ 
\K{decreases the geodesic distance but may decrease or increase the} Euclidean distance to the goal 
\K{at each time step}.
\K{For example, if the goal is behind the sofa, the agent must move around the sofa to reach it.}  \K{Importantly, the audio stream $A$ has complementary and potentially stronger information than $\Delta$ \Zcr{in} this regard.  Not only does the intensity of the audio source reflect the Euclidean distance to the target, but also the geometry of the room captured in the acoustics reveals geodesic distances.}  \KG{As we show in results, the visual and aural inputs are synergistic; neither fares as well on its own.}

\vspace*{-0.05in}
\paragraph{Implementation details.} The lengths of audio, visual, point vector, and final state, \ie, $L_A$, $L_V$, $L_{\Delta}$, and $L_S$ are $512$, $512$, $2$, and $1026$, respectively.
\Z{We use a single bidirectional GRU} with input size $512$, hidden size $512$, and we use one recurrent layer. 
We optimize \Zcr{the model} using Adam~\cite{KingmaICLR2015adam} with PyTorch defaults for coefficients for momentum and a learning rate of $2.5e-4$. We discount rewards with a decay of $0.99$. We train the network for $30M$ agent steps \cca{on Replica and $60M$ on Matterport3D}, which amounts to 105 and 210 GPU hours respectively.
\section{Experiments}
\label{sec:experiment}

Our main objectives
are to show:
\unnat{
\begin{enumerate}[label=\textbf{O.\arabic*},ref=O.\arabic*]\compresslist

    \item Tackling navigation with both sight and sound  (\ie, the proposed AudioPointGoal)
    leads to better navigation \K{and faster learning}. 
    \ZA{This demonstrates that audio has complementary information beyond merely goal coordinates that facilitates navigation}.\label{obj:audiopointgoal}

 \item Listening for an audio target in a 3D environment serves as a viable alternative to GPS-based cues. Not only does the proposed AudioGoal agent navigate better than the PointGoal agent, it does so without PointGoal's assumption of perfect odometry \Zcr{and even with noisy audio sensors}.  \KGcr{The AudioGoal task has the important advantage of realism: the agent autonomously senses the target in AudioGoal, whereas  
 the target is directly given to the agent via $\Delta$ in PointGoal---a rare scenario in real applications.}
    \label{obj:audiogoal}
    \item Audio-visual navigation can generalize to both new environments and new sound sources.  In particular, 
    audio-visual agents can navigate better with audio even when the sound sources are unfamiliar.
    \label{obj:audio-source}
\end{enumerate}
}

\begin{table}[b]
\vspace*{-0.2in}
\setlength{\tabcolsep}{2pt}
\centering
\caption{\small{Summary of \cn{SoundSpaces} dataset properties}}
\resizebox{0.95\linewidth}{!}{%
\begin{tabular}{c|c|c|c|c|c|c|c}
\toprule
Dataset         & \# Scenes & Resolution & Sampling Rate  & Avg. \# Node  & Avg. Area     & \# Training Episodes & \# Test Episodes\\
\midrule
Replica         & 18        & 0.5m       & 44100Hz  & 97            & 47.24  $m^2$  & 0.1M          & 1000 \\
Matterport3D    & 85        & 1m         & 16000Hz & 243           & 517.34 $m^2$  & 2M            & 1000 \\
\bottomrule
\end{tabular}%
}
\label{tab:dataset}
\end{table}

 \vspace*{-0.05in}
\paragraph{Datasets.}
\cca{Table~\ref{tab:dataset} summarizes \cn{SoundSpaces, which includes audio renderings for} the Replica and  Matterport3D datasets.}
\K{Each episode consists of a tuple: $\langle$scene, agent start location, agent start rotation, goal location, audio waveform$\rangle$.} 
\K{We generate episodes by choosing a scene and a random start and goal location.  To eliminate easier episodes,} we prune those that are either too short \K{(geodesic distance less than 4)} or can be completed by moving mostly in a straight line \K{(ratio of geodesic to Euclidean distance less than $1.1$)}. 
\cc{We ensure that at the onset of each episode the agent can hear the sound, since in some large environments the audio might be inaudible when the agent is very far from the sound source.}

 \vspace*{-0.05in}
\unnat{
\paragraph{Sound sources.}  Recall that the RIRs can be convolved with an arbitrary input waveform, which \KGcr{allows us} to vary the sounds across episodes.
We use \cca{102 copyright-free natural sounds of telephones, music, fans, and others (\url{http://www.freesound.org}).}  See Supp video for examples.  Unless otherwise specified, the sound source is the telephone ringing. 
\K{We stress that in all experiments, the environment (scene) at test time \KGcr{is unmapped and} has never been seen previously in training.  It is valid for sounds heard in training to also be heard at test time, e.g., a phone ringing in multiple environments will sound different depending on both the 3D space and the goal and agent positions.}   Experiments for \ref{obj:audio-source} examine the impact of varied train/test sounds.
}

 \vspace*{-0.05in}
\paragraph{Metrics.}
\K{We use the success rate normalized by inverse path length (SPL), the standard metric for navigation~\cite{anderson-eval}.} 
We  consider an episode successful only if the agent reaches the goal \emph{and} executes the \textit{Stop} action.

\begin{figure}[t]
\centering 
\includegraphics[width=0.9\textwidth]{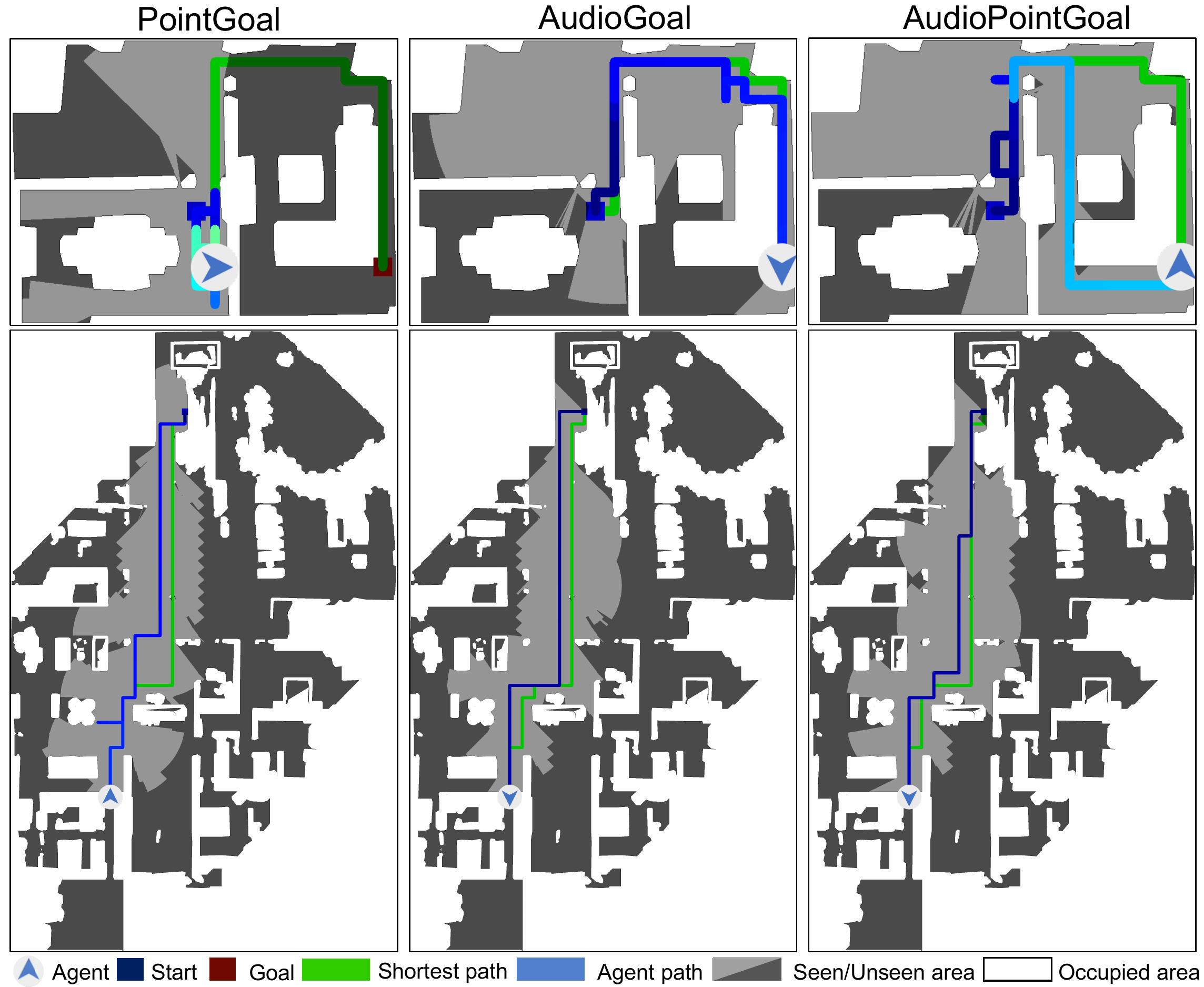}
\vspace*{-0.12in}
\caption{\small{\textbf{Navigation trajectories on top-down maps.} 
Agent path color fades from dark blue to light blue as time goes by. Green path indicates the shortest geodesic path.  
\KGcr{\textbf{Top:} Replica -} The PointGoal agent bumps into the wall several times trying to move towards the target, 
unable to figure out the target is actually located in another room.
In contrast, the AudioGoal and AudioPointGoal agents better sense the target: the sound travels through the door and the agent leaves the starting room immediately. \KGcr{\textbf{Bottom:} Matterport - the AudioGoal agent best avoids backtracking to efficiently reach the target in a large multi-room home.}
}}\vspace*{-0.15in}
\label{fig:trajectories}
\end{figure}

 \vspace*{-0.05in}
\paragraph{Baselines.}
We consider three non-learning baselines adapted from previous work \cite{habitat19iccv,chen_learning_2019}: \textsc{Random} chooses an action randomly among \{\emph{MoveForward}, \emph{TurnLeft}, \emph{TurnRight}\}. 
\textsc{Forward} always calls \emph{MoveForward} and if it hits an obstacle, it calls \emph{TurnRight} then resumes going forward and repeats.  \textsc{Goal follower} always first orients itself towards the goal and then calls \emph{MoveForward}. 
\KG{All three issue the \emph{Stop} action upon reaching the goal.}

\begin{table}[t]
\setlength{\tabcolsep}{5pt}
\centering\small
\caption{\small{Adding sound to sight and GPS sensing improves navigation performance significantly.  Values are \KG{success rate normalized by  path length} (SPL); higher is better.} }
\resizebox{\linewidth}{!}{
\begin{tabular}{c| c| c c | c c}
 \toprule
 \multicolumn{2}{c|}{} & \multicolumn{2}{c|}{Replica} & \multicolumn{2}{c}{Matterport3D} \\
 \multicolumn{2}{c|}{}        & PointGoal & AudioPointGoal & PointGoal & AudioPointGoal \\
 \midrule
 \multirow{3}{*}{Baselines}         & \textsc{Random}           & 0.044  & 0.044  &0.021 &0.021\\
                                    & \textsc{Forward}          & 0.063  & 0.063  &0.025 &0.025\\
                                    & \textsc{Goal follower}    & 0.124  & 0.124  &0.197 &0.197\\
 \midrule
 \multirow{3}{*}{Varying visual sensor}     & Blind  & 0.480     & \textbf{0.681}  &0.426 & \textbf{0.473}\\
                                            & RGB    & 0.521     & \textbf{0.632}  &0.466 & \textbf{0.521}\\
                                            & Depth  & 0.601     & \textbf{0.709}  &0.541 & \textbf{0.581}\\
 \bottomrule
\end{tabular}
}
\label{tab:pointgoal_vs_audiopointgoal}
\end{table}

\paragraph{\textbf{\ref{obj:audiopointgoal}: Does audio help navigation?}}

First we evaluate the impact of adding audio sensing to visual navigation by comparing PointGoal and AudioPointGoal agents. Table \ref{tab:pointgoal_vs_audiopointgoal} compares the navigation performance (in SPL) for both agents and the baselines on the test environments.  We consider three visual sensing capabilities: no visual input (Blind), raw RGB images, or depth images. (We found RGB+D was no better than depth alone.)

Audio improves accuracy significantly, showing the clear value in multi-modal perception for navigation.  Both learned agents do better with stronger visual inputs (depth being the strongest), though \KGcr{the margin between RGB and depth is a bit smaller for} AudioPointGoal.  
\KG{This is interesting because it suggests} that audio-visual learning captures geometric structure \KG{(like depth)} from the raw RGB images more easily than a model equipped with vision alone.  
As expected, the simple baselines perform poorly because they do not utilize any sensory inputs \KG{(and hence perform the same on both tasks).}

To see how audio influences navigation behavior,~\figref{fig:trajectories} shows example trajectories.  
See the Supp video for more.

\paragraph{\textbf{\ref{obj:audiogoal}: Can audio supplant GPS for an audio target?}}

Next we explore the extent to which audio supplies the spatial cues available from GPS sensing during (audio-)visual navigation.  This test requires comparing PointGoal to AudioGoal.  Recall that unlike (Audio)PointGoal, AudioGoal receives \emph{no} displacement vector pointing to the goal; it can only hear and see.

\begin{figure}[t]
\centering
\begin{tabular}{cc}
\hspace*{-0.2in}
    \subfigure[From perfect to noisy GPS]{\label{fig:noisy_pointgoal}
    \begin{tabular}{c}
        \includegraphics[width=0.42\textwidth]{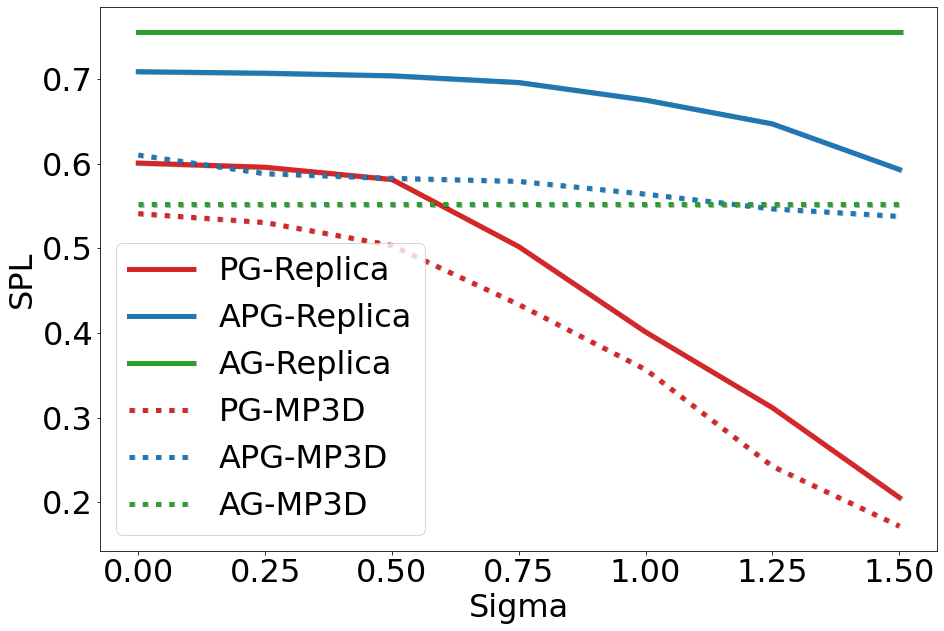}
    \end{tabular}}
    \subfigure[t-SNE of AudioGoal audio features]{\label{fig:tsne}
    \begin{tabular}{cc}
        \includegraphics[width=0.294\textwidth]{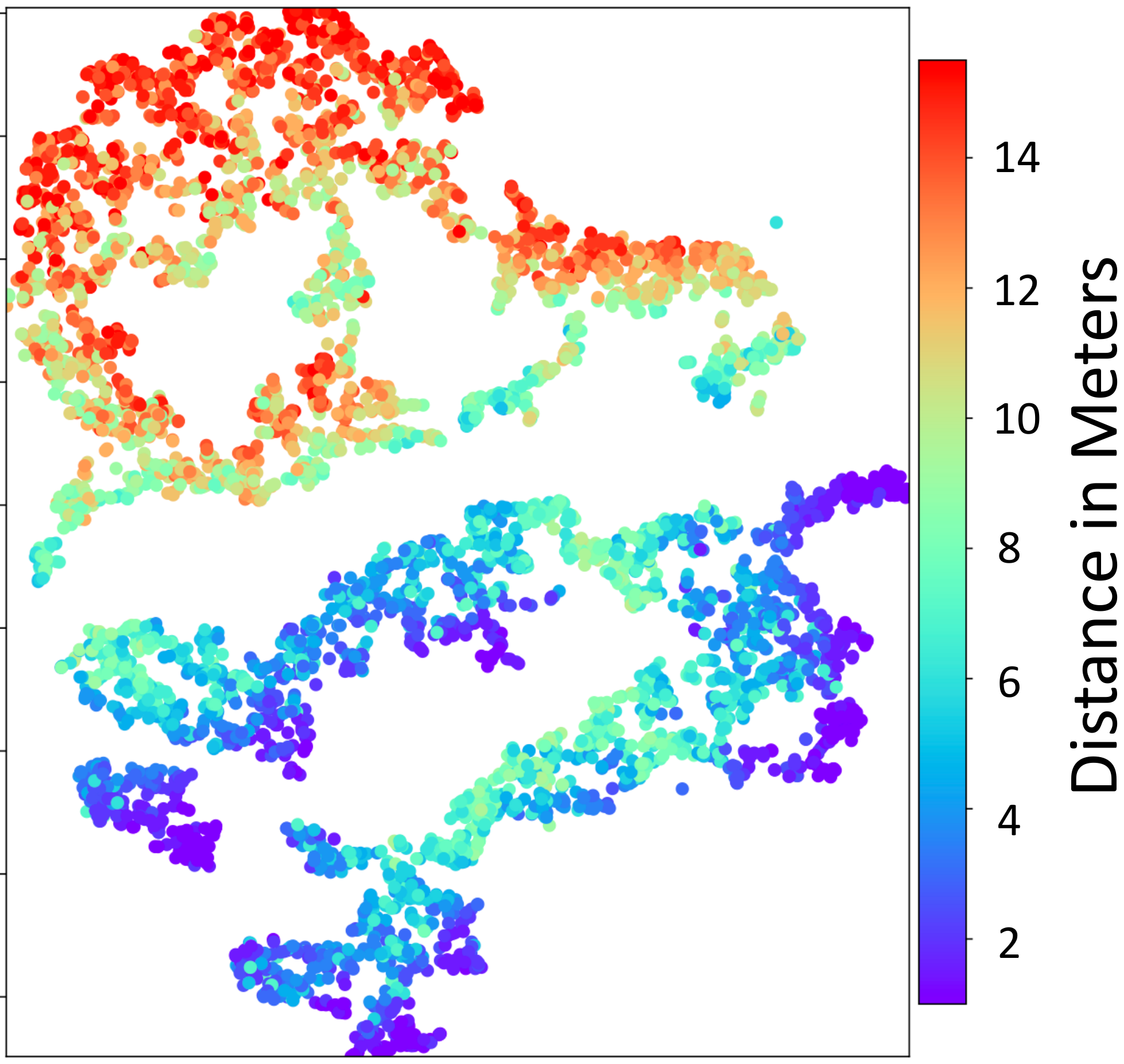}
        \includegraphics[width=0.294\textwidth]{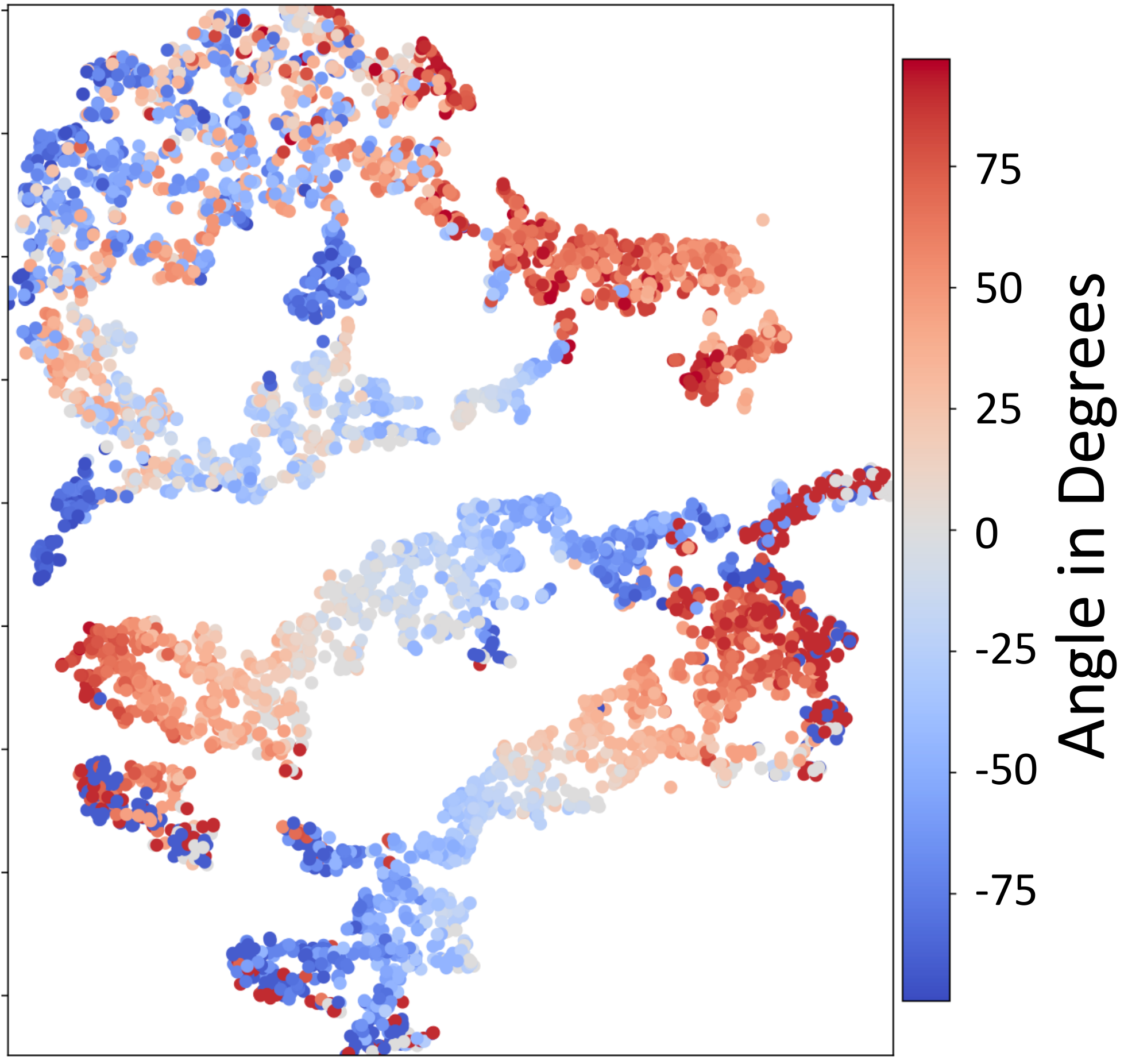}
    \end{tabular}}
\end{tabular}
    \vspace*{-0.15in}
\caption{\small{\textbf{Audio as a learned spatial sensor.} (a) Navigation accuracy with increasing GPS noise. Unlike existing PointGoal agents, our AudioGoal agent does not rely on GPS, and hence is immune to GPS noise.  
\KGcr{(b) t-SNE projection of audio features, 
color coded to reveal their correlation with the goal location (left) and direction (right),
\ie, source is far (red) or near (violet), and to the left (blue) or right (red) of the agent.}
}}
\vspace*{-0.15in}

\end{figure}

~\figref{fig:noisy_pointgoal} reports the navigation accuracy as a function of GPS quality.  The leftmost point uses perfect GPS that tells the PointGoal agents (but not the AudioGoal agent) the exact direction of the goal; for subsequent points, Gaussian noise of increasing variance is added, 
up to $\sigma=1.5$m. \KGcr{All agents use depth.}
While AudioGoal's accuracy is by definition independent of GPS failures, the others suffer noticeably.\footnote{\cca{Replica has \KGcr{more} multi-room trajectories, where audio gives clear cues of room entrances/exits \KGcr{(vs.~open floor plans in Matterport)}.
\KGcr{This may be why AG} is better than PG and APG  on Replica.}}
\KGcr{Furthermore, AudioPointGoal (APG) degrades much more gracefully than PointGoal (PG) in the face of GPS noise.}
This is evidence that \emph{the audio signal gives similar or even better spatial cues than the PointGoal displacements}---which are likely overly optimistic given the unreliability of GPS in practice \KG{and especially indoors.}  
 T-SNE~\cite{tsne} visualizations (Fig.~\ref{fig:tsne}) reinforce this finding: our learned audio features for AudioGoal naturally encode the distance and angle to the goal. 
 \KGcr{Note that these findings stand even with microphone noise: with 40dB SNR \cca{(bad microphone)}, SPL only drops marginally from 0.756 to 0.753 and from 0.552 to 0.550 on Replica and Matterport, respectively.}

\KG{Next we explore whether our AudioGoal agent learned \emph{more} than a pointer to the goal based on the sound intensity.  We run a variant of our model in which the audio input consists of only the intensity of the left and right waveforms; \cca{the audio CNN is removed,}
and the rest of the network in Fig~\ref{fig:model} remains the same.  This simplified audio input allows the agent to readily learn to follow the intensity gradient.  
The performance of the AudioGoal-Depth agent drops to an SPL of 0.291 and 0.014 showing that our model (SPL of 0.756 and 0.552 in Fig~\ref{fig:noisy_pointgoal}) does indeed learn additional environment information from the full spectrograms
to navigate more accurately.  See Supp.}

\begin{figure}[t]
\centering  
\subfigure{\includegraphics[width=\textwidth]{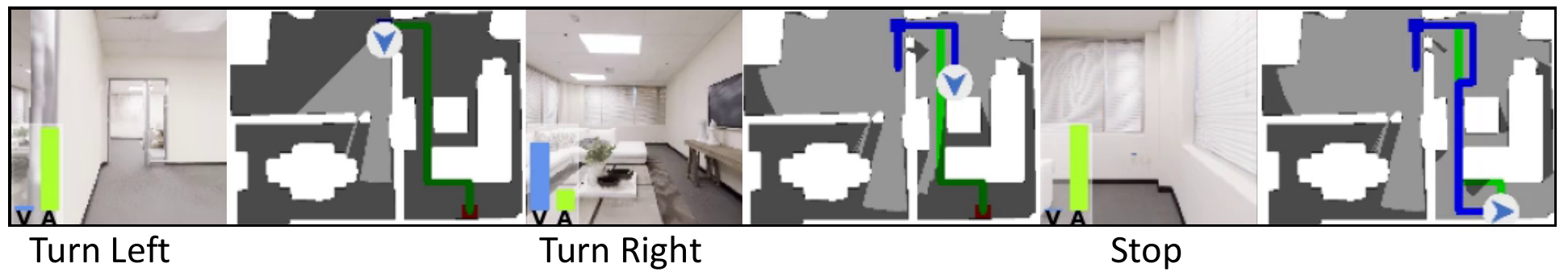}}\\ 
\subfigure{\includegraphics[width=\textwidth]{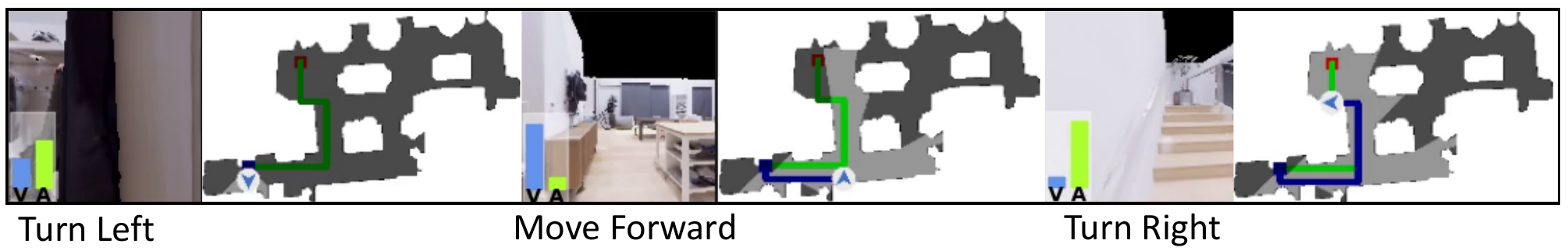}}\\
\vspace*{-0.15in}
\caption{\small{\KG{\textbf{Impact of each modality on action selection} for two AudioGoal episodes. 
We show one episode per row, and three sampled timesteps each.}  See Fig.~\ref{fig:trajectories} for legend.
Blue and green bars display the importance of vision and audio, respectively. \textbf{Top:} Initially, the agent relies on audio to tell that the goal is on its left and decides to turn left. Later, it uses vision to recognize \ccan{obstacles} in front of it and decides to \ccan{turn right}. Finally, the agent decides to stop because the sound intensity has peaked. 
\textbf{Bottom:} 
Initially, the agent decides to turn left, following the audio source.
\ccan{Then the agent uses vision to identify the free space and decides to move forward. Later, the agent relies more on audio to decide to turn right as it hears the target from the right. }
}}
\vspace*{-0.15in}
\label{fig:av_importance}
\end{figure}

\KG{We expect that the audio and visual input vary in their relative impact on the agent's decision making at any given time point, based on the environment context and goal placement.}  
To compute 
their impact, 
we \KG{ablate} 
each modality in turn by replacing it with its average training sample value, and compare the resulting action probability \KG{under our model} to that of the action chosen with both modalities. 
We calculate the importance of each input modality using the absolute difference of logarithmic action probability, normalized by the sum of the two ablations.
\KG{The greater the change in the selected action, the more impact that modality had on the learned agent's actual choice.} 
~\figref{fig:av_importance} \KG{and the Supp video} show examples of the AV impact scores alongside the egocentric view of the agent at different stages in the trajectory.   \KG{We see the agent draws dynamically on either or both modalities to inform its motions in the environment.}

\paragraph{\textbf{\ref{obj:audio-source}: What is the effect of different sound sources?}}

Next, we analyze the impact of the sound source. 
First, we explore generalization to novel sounds. \cca{We divide the 102 sound clips into 73/11/18 splits for train/val/test, respectively.  We train for AudioGoal (AG) and AudioPointGoal (APG), then validate and test on disjoint val and test sounds.}
In all cases, the test environments are unseen. 

Table~\ref{table:test} shows the results.  As we move left to right in the table, the sound generalization task gets harder: from a single heard sound, to variable heard sounds, to variable unheard sounds \Zcr{(see Supp for details on these three test settings)}. Note, the non-learning baselines are unaffected by changes to the audio and hence are omitted here.  Our APG agents \KGcr{almost} always outperform the PointGoal agent, even for unheard test sounds, strengthening the conclusions from Table~\ref{tab:pointgoal_vs_audiopointgoal}.  APG performs \KGcr{fairly} similarly on heard and unheard sounds, showing it has learned to balance all three modalities. 
On the other hand, AG's accuracy declines with varied heard sounds 
\KGcr{and unheard sounds.}  While it makes sense that the task of following an unfamiliar sound is harder, we also expect that larger training repositories of more sounds will resolve much of this decline.

\begin{table}[t]
\setlength{\tabcolsep}{6pt}
\centering\small
\caption{\small{Navigation performance (SPL) when generalizing to unheard sounds. Higher is better. Results are averaged over 7 test runs; all standard deviations are $\leq$ $0.01$.}}
\scalebox{0.85}{
\hspace*{-0.1in}
\begin{tabular}{c |c | c | c c | c c | c c}
\toprule
  & & &  \multicolumn{2}{ c| }{\textit{Same sound}} & \multicolumn{2}{ c |}{\textit{Varied heard sounds}} & \multicolumn{2}{ c }{\textit{Varied unheard sounds}}\\
Dataset & &       \emph{PG}&   \emph{AG}    & \emph{APG}  & \emph{AG}      & \emph{APG}  & \emph{AG}   & \emph{APG}     \\

\midrule
\multirow{3}{*}{Replica}    &Blind  & 0.480  & 0.673   & 0.681   & 0.449  & 0.633  & 0.277   & 0.649              \\
                            &RGB    & 0.521  & 0.626   & 0.632   & 0.624  & 0.606  & 0.339   & 0.562              \\
                            &Depth  & 0.601  & 0.756   & 0.709   & 0.645  & 0.724  & 0.454   & 0.707             \\
\midrule
\multirow{3}{*}{Matterport3D}       &Blind  & 0.426  & 0.438   & 0.473   & 0.352  & 0.500  & 0.278   & 0.497              \\
                            &RGB    & 0.466  & 0.479   & 0.521   & 0.422  & 0.480  & 0.314   & 0.448              \\
                            &Depth  & 0.541  & 0.552   & 0.581   & 0.448  & 0.570  & 0.338   & 0.538             \\
\bottomrule
\end{tabular}
}
\label{table:test}
\end{table}

\vspace*{-0.1in}
\section{Conclusion}
\vspace*{-0.1in}

We introduced the task of audio-visual navigation in complex 3D environments.  Generalizing a state-of-the-art deep RL navigation engine for this task, we presented encouraging results for audio's role in the visual navigation task.  \KG{The results show that when linked tightly to the egocentric visual observations, audio enriches not only the directional cues for a sound source, but also the spatial information about the environment---both of which our model successfully leverages for better navigation.}
\KG{Another important contribution of our work} is to enable audio rendering for Habitat with the publicly available Replica and Matterport3D environments, which can facilitate future work in the field.  Next we are interested in considering multi-agent scenarios, \KGcr{sim2real transfer}, moving sound-emitting targets, and navigating in the context of dynamic audio events.
\vspace*{-0.3in}
\section*{Acknowledgements}
\vspace*{-0.1in}
UT Austin is supported in part by DARPA Lifelong Learning Machines.  \KGcr{We thank Alexander Schwing, Dhruv Batra, Erik Wijmans, Oleksandr Maksymets, Ruohan Gao, and Svetlana Lazebnik for valuable discussions and support with the AI-Habitat platform.}

%
%
\clearpage
\bibliographystyle{splncs04}
\bibliography{unnat}

\clearpage
\section{Supplementary Material}
\label{sec:supplementary}

In this section we provide additional details about:
\begin{enumerate}
    \item Details of the audio simulation---including grid construction, mesh upgrades, acoustic simulation technique, and connectivity graph (as referenced in~Sec.~3 of the main paper).

    \item Additional illustrations of pressure fields from the audio simulation 
    and the sampled grid. 

    \item \KG{Reinforcement learning} training utilized in the network description (\KG{as referenced in Sec.~5 of the main paper.})

    \item Audio intensity baselines, as referenced in~Sec.~6 of the main paper.

    \item Heard/unheard sounds, as referenced in~Sec.~6,~Tab.~2, and~Tab.~3.

    \item Additional navigation trajectory examples, similar to~Fig.~4 \KG{in the main paper}.

\end{enumerate}

\subsection{Audio Simulation Details}
\mypara{Grid construction.}
We use an automatic point placement algorithm to determine the locations where the simulated sound sources and listeners are placed in a two-step procedure: adding points on a regular grid and then pruning. For adding points on a regular grid, first, we compute an axis-aligned 3D bounding box of a scene. Within this box we sample points from a regular 2D square grid with resolution 0.5m (Replica) or 1m (Matterport) that slices the bounding box in the horizontal plane at a distance of 1.5m from the floor (\K{representing the height of a humanoid robot}).

\KG{The second step prunes grid points in inaccessible locations.}
\K{To prune, we compute} how {\em closed} the region surrounding a particular point is.  This entails tracing $R$ uniformly-distributed random rays in all directions from the point, then letting them diffusely reflect through the scene up to $B$ bounces using a path tracing algorithm.
Simultaneously, \K{we compute} the total number of ``hits" $H$: the number of rays that intersect the scene. After all rays are traced, the {\em closed-ness} $C \in [0,1]$ of a point is given by $C = \frac{H}{R\cdot B}$.
A point is \K{declared} outside the scene if \K{$C < C_{min}$}.
 the value of $C$ for a particular point is below a threshold $C_{min}$.
Finally, we \K{remove} points that are within a certain distance $d_{min}$ from the nearest geometry,
\K{as identified} using the shortest length of the initial rays traced from the point in the previous pruning step.

For all scenes we use $R = 1000$, $B = 10$ and $d_{min} = 5$cm. This value of $d_{min}$ was chosen to avoid placement of points inside walls or in small inaccessible areas.
We find $C_{min} = 0.5$ works for most scenes. The exceptions are scenes with open patio areas, where we found $C_{min} = 0.1$ works best to provide a sufficient number of points on the patio.

\mypara{\KG{Materials and transmission model.}}
In addition to its geometry, a room's \emph{materials} affect the RIR, \KG{as discussed in the main paper.}
To capture this aspect, we use the semantic \K{labels} provided in Replica to determine the acoustic material properties of the geometry. For each semantic class that was deemed to be acoustically relevant, we provide a mapping to an equivalent acoustic material from an existing material database \cite{egan1989}. For the {\em floor}, {\em wall}, and {\em ceiling} classes, we assume acoustic materials of carpet, gypsum board, and acoustic tile, respectively. \cc{This helps simulate \K{more realistic} sounds than if a single material \K{were} assumed for all surfaces.}
\cc{In addition, we add a ceiling to those Replica scenes that lack one, which is necessary to simulate the acoustics accurately.}

The simulation also includes a path-tracing simulation through walls according to their material properties.  Each material has absorption, scattering, and transmission coefficients.  We use a transmission model similar to that used in graphics rendering. While this is modeled to ensure precision of the simulation, the impact of transmission is generally small compared to the propagation of sound through open doors \cite{locher_differences_2018}.

\mypara{Acoustic simulation technique.}
During the simulations, \K{we compute} the room impulse responses between all pairs of points, producing $N^2$ RIRs. The simulation technique stems from the theory of geometric acoustics (GA), which \K{supposes} 
sound can be treated as a particle or ray rather than a wave~\cite{savioja2015}. This class of simulation methods is capable of accurately predicting the behavior of sound at high frequencies, but requires special modeling of wave phenomena (e.g., diffraction) that occur at lower frequencies.
Specifically, our acoustic simulation is based on a bidirectional path tracing algorithm~\cite{veach1995}  modified for room acoustics applications~\cite{cao2016}.
Additionally, it uses a recursive formulation of multiple importance sampling (MIS) to improve the convergence of the simulation \cite{georgiev2012}.

The simulation begins by tracing rays from each source location
\K{in $\mathcal{S}$.}
These source rays are propagated through the scene up to a maximum number of bounces (\ZA{$200$}).
At each ray-scene intersection of a source path, information about the intersected geometry, incoming and outgoing ray directions, and probabilities are cached.
After all source rays are traced, the simulation traces rays from a listener location \K{in $\mathcal{L}$}.
These rays are again propagated through the scene up to a maximum number of bounces.
At each ray-scene intersection of a listener path, rays are traced to connect the current path vertex to the path vertices previously generated from all sources.
If a connection ray is not blocked by scene geometry, a path from the source to listener has been found.
The energy throughput along that path is multiplied by a MIS weight and is accumulated to the impulse response for that source-listener pair.
After all rays have been traced, the simulation is finished. 

\K{We perform} the simulation in parallel for four logarithmically-distributed frequency bands.\footnote{\K{[0Hz,176Hz], [176Hz,775Hz], [775Hz,3409Hz], [3409Hz,20kHz]}} These bands cover the human hearing range and are uniform in their distribution from a perceptual standpoint.
For each band, the simulation output is a histogram of sound energy with respect to propagation delay time at audio sample rate (44.1kHz for Replica and 16kHz for Matterport).
Spatial information is also accumulated in the form of low-order spherical harmonics for each histogram bin.
After ray tracing, these energy histograms are converted to pressure IR envelopes by applying the square root, and the envelopes are multiplied by bandpass-filtered white noise and summed to generate the frequency-dependent reverberant part of the monaural room impulse response \cite{kuttruff1993}.

Ambisonic signals (\K{roughly speaking, the audio equivalent of a 360$^\circ$ image}) are generated by decomposing a sound field into a set of spherical harmonic basis.
We generate ambisonics by multiplying the monaural RIR by the spherical harmonic coefficients for each time sample.
Early reflections (ER, paths of order $\le 2$) are handled specially to ensure they are properly reproduced.
ER are not accumulated to the main energy histogram, but are instead clustered together based on the plane equation of the geometry involved in the reflection(s).
Then, each ER cluster is added to the final pressure IR with frequency-dependent filtering corresponding to the ER energy and its spherical harmonic coefficients.

The result of this process is 2nd-order ambisonic pressure impulse responses that can be convolved with \K{arbitrary new}  monaural source audios to generate the ambisonic audio heard at a particular listener location.    \K{We  convert the ambisonics to binaural audio~\cite{zaunschirm2018binaural} in order to represent an agent with two human-like ears, for whom perceived sound depends on the body's relative orientation in the scene.}

\subsection{Visualizing Audio Simulations}
Next we illustrate the pressure field visualization of two other scenes in the Replica dataset. In~\figref{fig:pressure_apt2}, we display another big scene (apartment\_2) with four rooms, with the audio source inside one of the rooms. Notice how the pressure decreases from the source along geodesic paths, which leads to doors serving as secondary sources or intermediate goals that lead the agent in the right direction. 

\begin{figure}[t]\label{fig:pressure_apt2}
    \centering
    \includegraphics[trim=0 0 0 0 ,clip,width=0.5\textwidth]{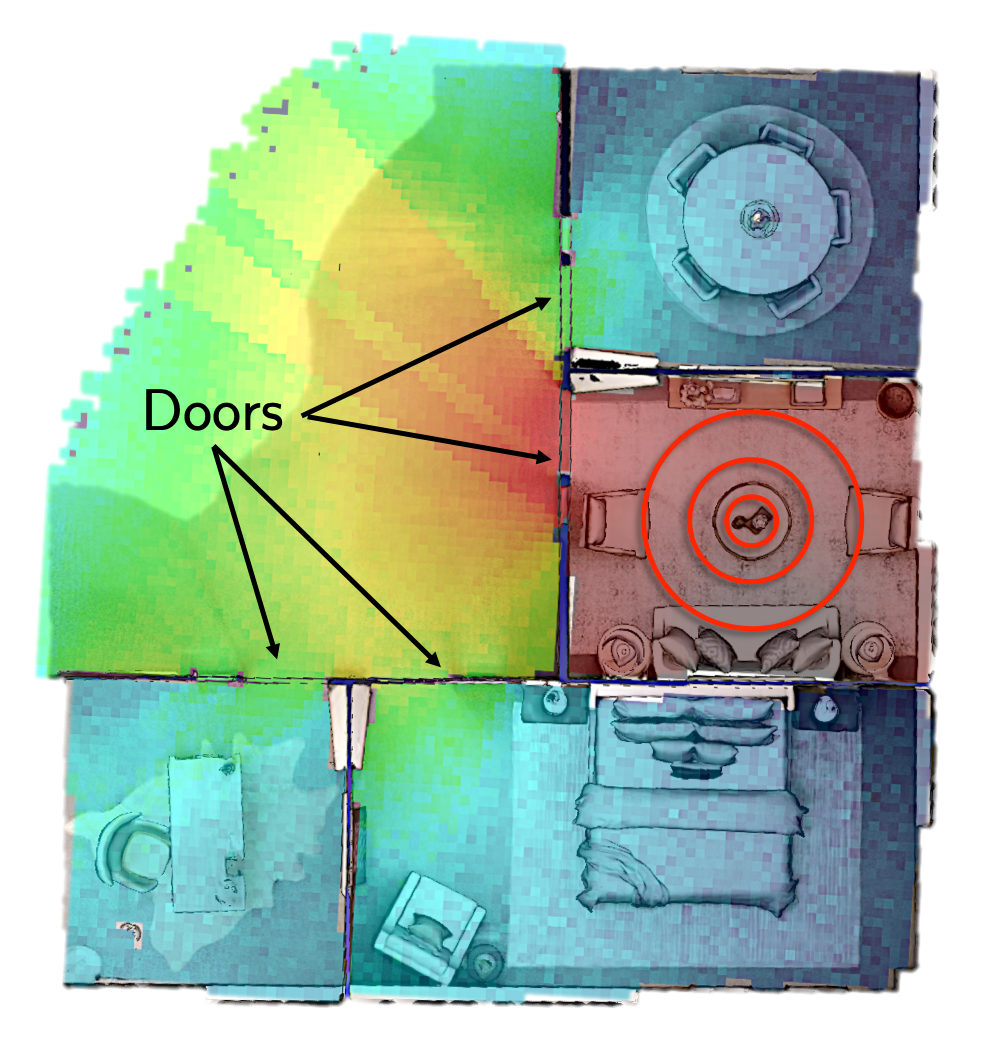}
    \caption{\textbf{Pressure field of audio simulation} overlaid on the top-down map of apartment\_2 from Replica~\cite{straub2019replica}. Our audio-enabled agent gets rich directional information about the goal, since the pressure field variation is correlated with the shortest distance. Notice the discontinuities across walls and the gradient of the field along the \emph{geodesic} path an agent must use to reach the goal (different from shortest Euclidean path). As a result, to an agent standing in the top right or bottom rooms, the audio reveals the door as a good intermediate goal.  In other words, the audio stream signals to the agent that it must leave the current room to get to the target.  In contrast, the GPS displacement vector would point through the wall and to the goal, which is a path the agent would discover it cannot traverse.  \KG{Note that the visual stream is essential to couple with the audio stream in order to navigate around obstacles.}
    }
    \label{fig:pressure_apt2}
\end{figure}

\begin{figure}[t]
    \centering
    \includegraphics[trim=0 0 0 0 ,clip,width=0.5\linewidth]{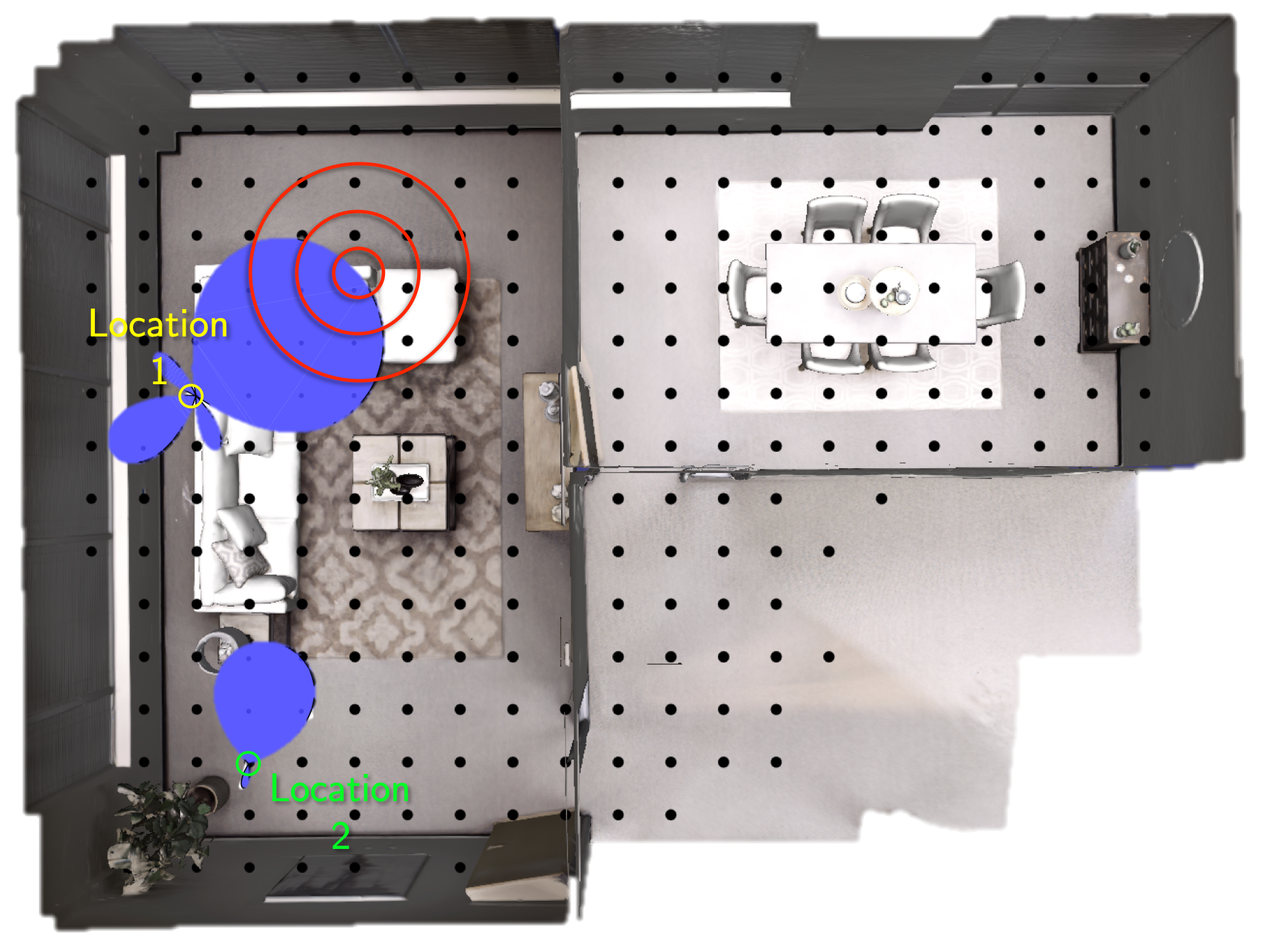}
    \caption{\textbf{Visualizing ambisonics.} We visualize the ambisonics components (blue \textit{lobes}) of the impulse response. Notice that the ambisonics sound fields characterize direction and intensity of the incoming energy.}\label{fig:audio_lobes}
\end{figure}

\KG{\figref{fig:audio_lobes} displays} a second-order ambisonics representation showing the direction and intensity of the incoming direct sound. Particularly, it demonstrates the spatial properties of the audio simulation at two receiver locations. Recall that we render impulse responses for source and receiver positions sampled from a grid in each scene. These impulse responses are stored in ambisonics and converted to binaural to mimic the signals received by a human at the entrance of the ear canal. We create~\figref{fig:audio_lobes} by evaluating the incoming energy of the direct sound (excluding reflections and reverberation) at the horizontal plane.\footnote{The minor side lobes pointing in directions other than the source are a result of representing the sound field as a $2^{nd}$ order ambisonics signal, thus using only 9 spherical harmonics. We refer the reader to~\cite{daniel2003spatial,Rafaely2015,Zotter2019} for more details on ambisonics sound field representation. 
}
The greater the energy the bigger the size, and the orientation depicts the angular distribution of energy. In Location 1 energy comes predominantly from its right.  Since it is closer to the audio source, the directional sound field has more energy than Location 2.

\subsection{Reinforcement Learning Training Details}
In the following, we provide details of our reinforcement
learning (RL) formulation for navigation tasks. This notation links to~Sec.~4 and~Fig.~3 in the main paper.

An agent embedded in an environment must take actions from an action space $\cA$ 
to accomplish an end goal. For our tasks, the actions are navigation motions: $\cA=\{MoveForward, TurnLeft, TurnRight, Stop\}$. At every time step
 $t=\{0,1,2,\ldots,T-1\}$ the environment is in some state $s_t\in\mathcal{S}$, but the agent obtains only a partial
observation of it in the form of $o_t$. Here $T$ is a maximal time horizon, which corresponds to 500 actions for our task. The observation $o_t$ is a combination of the audio, visual, and displacement vector inputs.

Using information about the previous time steps $h_{t-1}$ and current observation
$o_t$, the agent develops a policy
$\pi_{t,\theta}:\cA\to[0,1]$, where $\pi_{t,\theta}(a|o_t,h_{t-1})$ is the probability that the agent chooses to take action $a\in\cA$ at time $t$. We use the shorthand of $\pi_{t,\theta}(o_t,h_{t-1})$ to show the feed-forward nature of the actor head. After the agent acts, the environment goes into a new state $s_{t+1}$ and the agent receives individual rewards
$r_{t}\in\mathbb{R}$.  

The agent optimizes its \textit{return}, \ie the expected discounted, cumulative rewards
\begin{equation}
G_{\gamma,t} = \sum_{t=0}^{T-1}\gamma^tr_t,
\end{equation}
where $\gamma\in[0,1]$ is the discount factor to modulate the emphasis on recent or long term rewards. The value function $V_{t,\theta}(o_t, h_{t-1})$ is the expected return. 
The particular reinforcement learning objective we optimize directly follows from Proximal Policy Optimization.  We refer the readers to~\cite{ppo} for additional details on optimization.  

\subsection{Audio Intensity Baseline}

\KG{In the main paper, we presented an audio intensity baseline in Sec~\ref{sec:experiment}.  It is an ablation of our model where the policy is learned directly from the intensity of the left and right waveforms together with the depth-based visual stream.} We compute the intensity of audio using the Root-Mean-Square (RMS) of channel's waveform, which produces two real numbers as the audio feature.  \KG{We showed that it is inferior to our approach, meaning that our model is able to learn additional environment information from the full spectrograms.  Here we provide the parallel results for the blind and RGB visual streams (\tabref{tab:intensity}).}

\Z{We see a significant drop in performance when using audio intensity only compared to spectrograms. This \KG{demonstrates} that our model \KG{extracts} useful acoustic features for navigation (\eg relative angle to goal, \KG{major obstacles}) that go beyond just intensity.}

\setlength{\tabcolsep}{15pt}
\begin{table}[t]    
    \centering
    \caption{Intensity \Z{only versus spectrograms as audio input for our model and} with different visual inputs for AudioGoal agents (blind / RGB / depth).}
    \begin{tabular}{c|c|c }
    \toprule
    \Z{Audio Features} & Replica & MP3D \\
    \midrule
    \Z{Intensity only} & 0.276 / 0.177 / 0.291 &0.173  / 0.003  / 0.014\\
    \midrule
    \Z{Spectrograms} & 0.673 / 0.626 / 0.756 &0.438  / 0.479  / 0.552\\
    \bottomrule
    \end{tabular}%
    \label{tab:intensity}
\end{table}

\subsection{Heard/Unheard Dataset Splits}
In the following we provide details about the sounds used in~Sec.~6. 
\cca{We utilize 102 copy-free natural sounds across a wide variety of categories: air conditioner, bell, door opening, music, computer beeps, fan, people speaking, telephone, and etc. We divide these 102 sounds in to non-overlapping 73/11/18 splits for train, validation and test. 

For~Tab.~2 and the \textit{same sound} experiment in~Tab.~3 of the main paper, we use the sound source of 'telephone'.
In~Tab.~3, for the \textit{varied heard sounds} experiment we train using the 78 sounds and test on unseen scenes with the same sounds. Recall that the audio observations vary not only according to the audio file but also the 3D environment.
For the \textit{varied unheard sounds} experiment, we use the 78 sounds for training scenes, and generalize to unseen scenes as well as unheard sounds. Particularly, we utilize the 11 sounds for validation scenes, and the remaining 18 sounds for test scenes.
}

\subsection{Additional Navigation Trajectory Examples}
\begin{figure}[h]
\centering     
\includegraphics[width=\textwidth]{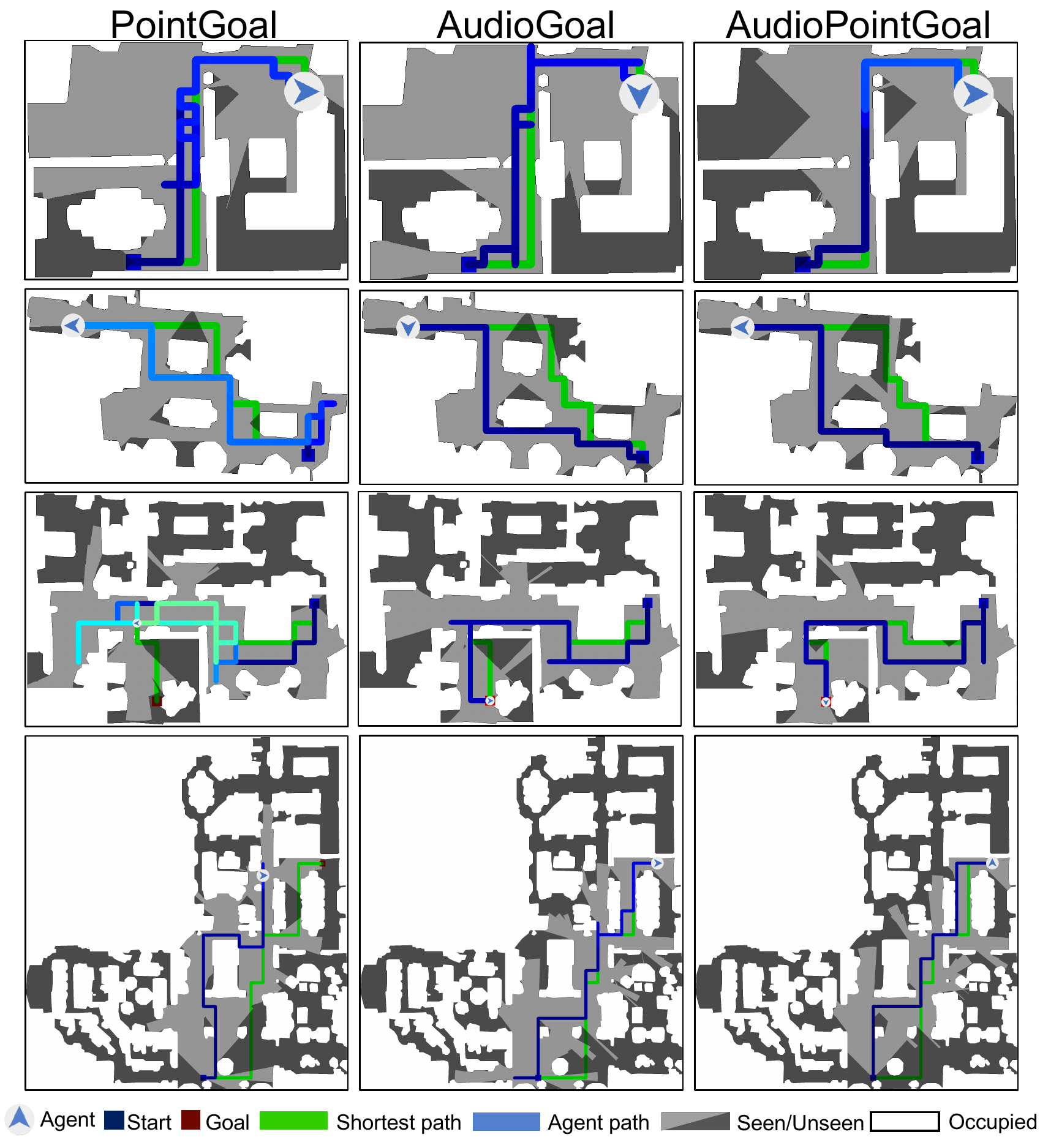}
\caption{\textbf{Navigation trajectories on top-down maps.}
\cca{The top two and bottom two rows are environments in Replica and Matterport3D, respectively.}
Agent path color fades from dark blue to light blue as time goes by. Green path indicates the shortest geodesic path. 
In this figure, we show navigation trajectories of three agents in varied test environments. 
The AudioGoal agent and AudioPointGoal agent navigate more efficiently compared to PointGoal agent.
Best viewed in color.  
}
\label{fig:trajectories2}
\end{figure}

\figref{fig:trajectories2} shows four additional trajectory examples of three agents in different test environments \cca{of Replica and Matterport3D.
These trajectories show the AudioGoal agent and AudioPointGoal agent navigate to goals more efficiently compared to PointGoal. 
}

\end{document}